\documentclass{article}

\usepackage{microtype}
\usepackage{graphicx}
\usepackage{subcaption}
\usepackage{booktabs} 
\usepackage{xcolor}
\usepackage{graphicx}
\usepackage{subcaption}
\usepackage{listings}

\usepackage{hyperref}

\usepackage[accepted]{icml2026}

\usepackage{amsmath}
\usepackage{amssymb}
\usepackage{mathtools}
\usepackage{amsthm}

\usepackage[capitalize,noabbrev]{cleveref}

\theoremstyle{plain}

\theoremstyle{definition}

\theoremstyle{remark}

\usepackage[textsize=tiny]{todonotes}

\definecolor{avgreen}{HTML}{93C47D}   
\definecolor{avred}{HTML}{EA9999}     
\definecolor{XinyiColor}{rgb}{0.37,0.62,0.63}
\definecolor{jadencolor}{RGB}{120,40,130}

\icmltitlerunning{DocHop: Benchmarking Out-of-domain Multi-hop Reasoning in Information-Dense Documents}

\begin{document}

\twocolumn[
  \icmltitle{DocHop: Benchmarking Out-of-domain Multi-hop Reasoning in Information-Dense Documents}



  \icmlsetsymbol{equal}{*}

  \begin{icmlauthorlist}
    \icmlauthor{Zhuoran Yu}{yyy}
    \icmlauthor{Le Thien Phuc Nguyen}{yyy}
    \icmlauthor{Jaden Park}{yyy}
    \icmlauthor{Xinyi Gu}{mmm}
    \icmlauthor{Zexue He}{sss}
    \icmlauthor{Soochahn Lee}{kkk}
    \icmlauthor{Rogerio Feris}{iii}
    \icmlauthor{Yong Jae Lee}{yyy}
  \end{icmlauthorlist}

  \icmlaffiliation{yyy}{University of Wisconsin-Madison}
  \icmlaffiliation{mmm}{Massachusetts Institute of Technology}
  \icmlaffiliation{sss}{Stanford University}
  \icmlaffiliation{kkk}{Kookmin University}
  \icmlaffiliation{iii}{MIT-IBM Watson AI Lab, IBM Research}

  \icmlcorrespondingauthor{Zhuoran Yu}{zhuoran.yu@wisc.edu}

  \icmlkeywords{Machine Learning, ICML}

  \vskip 0.3in
]



\printAffiliationsAndNotice{}  

\begin{abstract}
Multimodal Large Language Models (MLLMs) have achieved strong performance on structured visual understanding tasks such as chart and document question answering. However, existing benchmarks typically evaluate these domains in isolation, leaving underexplored a key capability: whether models can use textual context to determine how chart evidence should be selected, interpreted, and aggregated.
We introduce \textsc{DocHop}, a benchmark for integrated chart--context reasoning in document-style images. 
In \textsc{DocHop}, the document narrative specifies multi-step compositional constraints, while charts provide the corresponding data values. Questions are grounded on a semantic reference label defined in the narrative, requiring models to resolve target entities from context before aggregating evidence across multiple charts. 
To enable systematic evaluation, we construct \textsc{DocHop} via a stochastic logic-first generation pipeline with controllable reasoning depth and visual density, covering 2,074 examples across six task categories. 
Experiments on a wide range of proprietary and open-source MLLMs show a substantial gap to human performance: annotators achieve over 90\% accuracy, while the best model reaches only 62.83\%. Reasoning-enhanced models consistently show improved results, but performance degrades as reasoning complexity increases. 
Overall, \textsc{DocHop} provides a controlled testbed for challenging multi-hop document reasoning.
\end{abstract}

\section{Introduction}
Recent Multimodal Large Language Models (MLLMs)~\cite{GPT5, liu2024llavanext, Qwen3-VL, bai2025qwen2, Gemini2.5, liu2023visual, dai2023instructblip} have shown promising results in structured image understanding, spanning domains such as charts~\cite{chartqa, ChartQAPro, chartbench, charxiv, ChartX}, documents~\cite{mathew2021docvqa}, and webpage-based interfaces~\cite{liu2024visualwebbench, liu2024harnessing}. However, existing benchmarks predominantly evaluate these domains in isolation. For example, chart-centric and document-centric datasets are typically constructed and tested separately, each focusing on a particular form of visual information. This separation overlooks a key aspect of real-world presentation: the interplay between visual data and textual context. In many practical scenarios, charts and text function together, where the surrounding narrative provides definitions, contextual grounding, or reasoning instructions that are essential for correctly interpreting the accompanying visual evidence. These scenarios highlight an important but underexplored capability: models must bind textual constraints to visual evidence and then use the grounded evidence for downstream reasoning.

In this paper, we introduce \textsc{DocHop}\footnote{\url{https://www.zhuoranyu.com/dochop-page/}}, a benchmark designed to evaluate MLLMs on integrated document understanding that combines chart-based evidence with surrounding narrative context. The core intuition is to separate reasoning logic from numerical evidence: the document context specifies the multi-step constraints, while the charts provide the corresponding data values. Models must jointly interpret the document narrative and the chart values, using the described reasoning trace to locate and aggregate visual evidence across multiple charts. This design enables a rigorous evaluation of visual-intensive multi-hop reasoning, rather than simple single-hop lookup. 
In the following paragraphs, we highlight the key properties of \textsc{DocHop}.

\begin{figure*}[t]
    \centering
    \includegraphics[width=\textwidth]{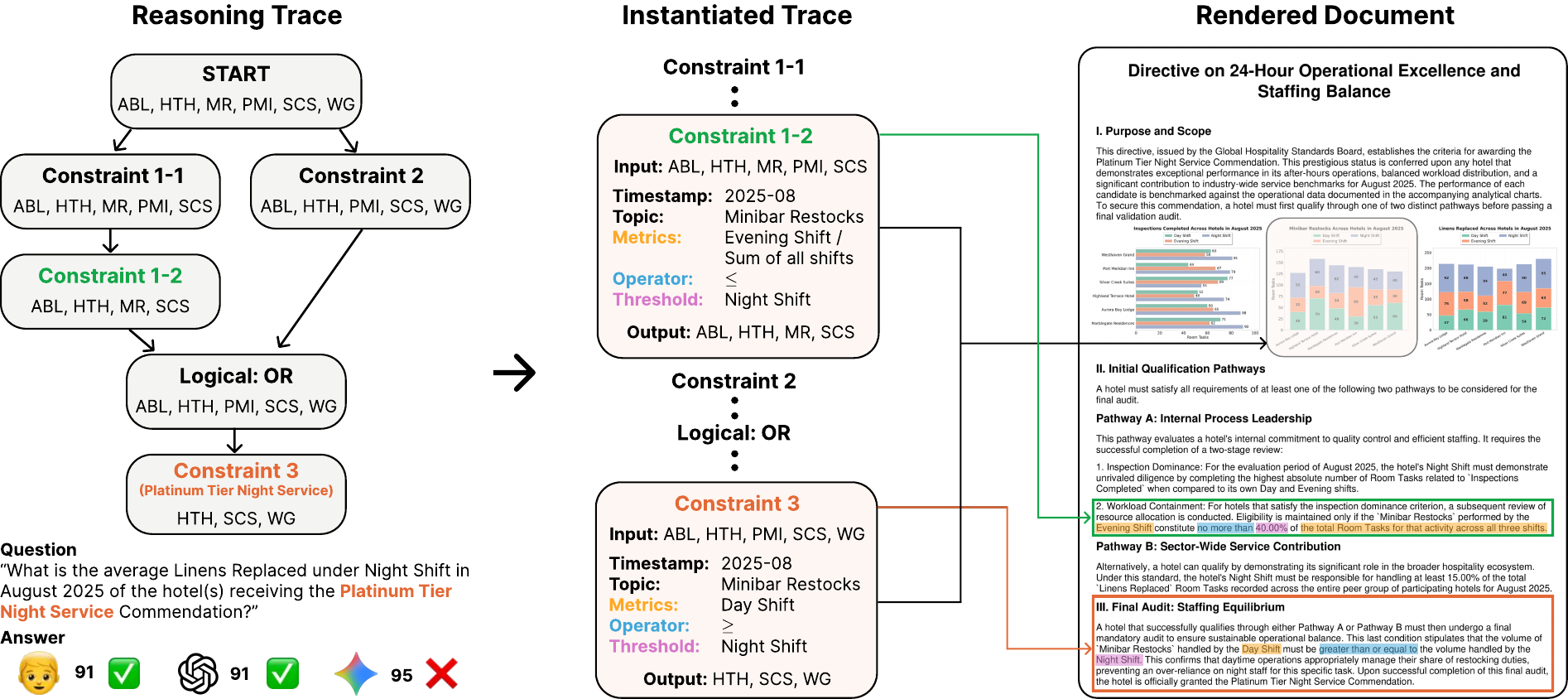}
    \caption{\textbf{Illustration of an example in DocHop}. 
Each document instance corresponds to a stochastically-sampled multi-hop reasoning trace, which is transformed into a narrative context interleaved with multiple charts. The narrative specifies the compositional constraints of the trace and concludes by assigning a semantic reference label to the entities implied by its execution. Questions are then grounded on this label rather than explicit entity names, requiring models to resolve targets from the document context and aggregate numerical evidence from the charts. The colored arrows indicate where the constraints are presented in the document; the black arrow indicates the specific chart the model needs to look at to check if the input entities satisfy the constraint.}
    \label{fig:teaser}
\end{figure*}

\textbf{Integrated Chart-Context Reasoning}.
\textsc{DocHop} consists of document-style images in which multiple charts are interleaved with explanatory narrative context. The narrative does not merely provide background text; it specifies the multi-step reasoning constraints that govern how candidate entities should be selected across charts. Crucially, it concludes by assigning a semantic reference label to the entities obtained by executing this reasoning trace under its logical composition. Questions are then formulated to refer to this label rather than explicit entity names, requiring models to first resolve the target entities from the document context and subsequently retrieve and aggregate the corresponding numerical evidence from the charts. This design ensures that correct answers depend on joint understanding of both the narrative constraints and the visual data. An illustrative example can be found in Figure~\ref{fig:teaser}.

\textbf{Controllable Reasoning Complexity}.
To achieve a rigorous evaluation, we employ a stochastic logic-generation pipeline rather than relying on noisy web-scraped data. For each instance, we first sample a reasoning trace---a structured chain of conditional checks that necessitates aggregating information from multiple distinct charts. We then prompt an LLM to generate both the underlying numerical data and the accompanying textual document to strictly satisfy this pre-defined trace. This \textit{reverse-engineering} approach ensures semantic consistency between the text and the charts. Crucially, it allows us to explicitly control difficulty along two axes: reasoning depth (the number of logical hops) and visual density (the number of charts). This results in a diverse difficulty distribution, enabling a systematic diagnostic of model limitations across the entire complexity spectrum---from lower-depth chart--context reasoning to high-load multi-step aggregation---as detailed in Section~\ref{sec:dochop}.

\textbf{Diverse Reasoning Contexts}. Our stochastic generation process ensures high variance across logical, visual, and thematic dimensions. The reasoning traces are randomly generated to produce varied structures, ranging from linear chains to multi-branch compositions. Together, these variations broaden the coverage of chart--context reasoning patterns evaluated by the benchmark. Across semantic domains, the benchmark spans 480 distinct topics, and its documents include 7 different chart types. In total, DocHop comprises 2,074 examples covering 6 distinct tasks: Value Retrieval, Counting, Numeric Reasoning, Ranking, Hypothetical Reasoning, and Fact Checking. These are instantiated via 41 unique question templates, ensuring a comprehensive evaluation of model generalizability.

\textbf{Multi-Stage Quality Verification}. 
To ensure reliability, we implement a multi-stage verification protocol. First, leveraging the structured nature of our generation pipeline, we programmatically validate the numerical data in every example. Since the underlying data tables and reasoning traces are fully accessible, we execute the trace against the data tables to deterministically verify that the generated chart values satisfy the instantiated constraints and that the ground-truth answers are mathematically correct. Second, we conduct human verification of the document context to ensure that the generated narrative faithfully verbalizes the underlying reasoning trace, including the textual constraints, logical compositions, and semantic reference label. Third, we manually inspect the rendered document images to ensure visual readability and completeness, checking for issues such as truncated context, overlapping layout elements, occluded charts, or illegible labels. This combination of programmatic validation and human review supports high-fidelity evaluation standards.

Since \textsc{DocHop} is synthetically constructed rather than sampled from naturally occurring documents, we position it as an out-of-domain multi-hop reasoning benchmark for chart--document integration. Its goal is not to model a specific real-world document distribution, but to test whether models can use narrative context to identify the relevant chart evidence and aggregate it across multiple charts.

We evaluate DocHop on a wide range of proprietary and open-source MLLMs. The results show that DocHop remains challenging but not intractable: human annotators achieve over 90\% accuracy, while the best-performing model (GPT-5.2 Reasoning~\cite{openai2025gpt52}) reaches 62.83\% overall. Reasoning-enhanced variants consistently outperform their non-reasoning counterparts across both GPT and Gemini families. Proprietary models remain stronger overall, with open-source baselines typically demonstrating 9--24\% accuracy in our evaluation. 

Moreover, DocHop enables controlled analysis along reasoning depth and chart-count axes, where performance degrades steadily as either complexity factor increases (Figure~\ref{fig:reasoning_performance}). Finally, we provide qualitative failure case studies to better understand common breakdowns in narrative grounding and cross-chart evidence aggregation.

\section{Related Work}
\label{sec:related}

\subsection{Multimodal Large Language Models}

Recent Multimodal Large Language Models (MLLMs) extend large language models with visual encoders to jointly process images and text. Through large-scale pretraining and instruction tuning, both proprietary and open-source MLLMs have demonstrated strong performance across a wide range of multimodal understanding tasks. Representative proprietary systems include the GPT series~\citep{GPT5}, Gemini~\citep{Gemini2.5}, and Claude models~\citep{anthropic2025claudehaiku45, anthropic2025claudesonnet45}. In parallel, open-source models such as LLaVA-style architectures~\citep{liu2024llavanext, liu2023visual}, IDEFICS~\citep{laurenccon2024matters, laurençon2024building}, Qwen-VL~\citep{bai2025qwen2, Qwen3-VL}, Molmo~\cite{deitke2025molmo}, and InternVL~\cite{internvl35} provide competitive performance and have enabled reproducible research across diverse benchmarks. These general-purpose MLLMs adopt unified architectures that integrate visual perception and language reasoning, enabling zero-shot and few-shot generalization across tasks involving images, documents, and structured layouts. As a result, they have become standard baselines for evaluating multimodal understanding capabilities in contemporary benchmarks. However, their capabilities in complex multi-modal multi-hop reasoning tasks remain underexplored.

\subsection{Benchmarks for Structured Multimodal Reasoning}

A large body of benchmarks has been proposed to evaluate visual reasoning over structured images---particularly in charts, documents, and webpages. Early chart question answering datasets such as FigureQA~\citep{figureqa}, DVQA~\citep{dvqa}, LEAF-QA~\citep{leafqa}
, and PlotQA~\citep{plotQA} rely heavily on questions generated from limited templates. 

ChartQA~\citep{chartqa} introduces human-authored questions on real-world charts and improves linguistic diversity, but still focuses on single-chart reasoning without requiring external textual context. More recent benchmarks on chart and document question answering, including ChartBench~\citep{chartbench}, MMC~\citep{MMC}, ChartX~\citep{ChartX}, ChartXiv~\citep{charxiv}, ChartMuseum~\citep{ChartMuseum}, ChartQAPro~\citep{ChartQAPro}, and DocVQA~\citep{mathew2021docvqa} increase image diversity and question difficulty by incorporating scientific figures, multi-chart inputs, or open-ended answers. Recently, DocHop-QA~\citep{park2025dochop} studies multimodal multi-hop question answering over document collections. However, these benchmarks largely follow a single-hop or limited multi-chart paradigm, where the reasoning logic is implicit in the question rather than explicitly provided and followed across modalities. In contrast to prior benchmarks, DocHop is designed to evaluate multi-round cross-modal reasoning within a single document. 
\begin{figure*}[ht]
    \centering
    \includegraphics[width=\textwidth]{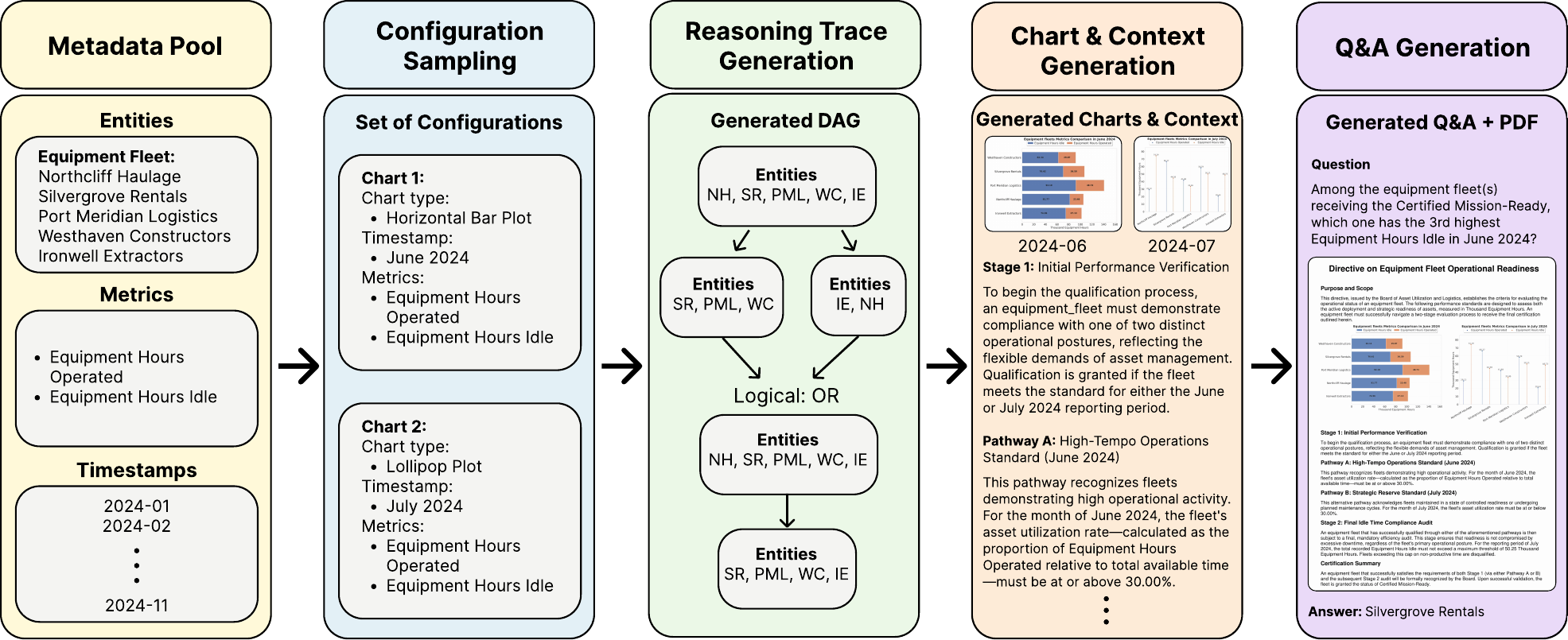}
    \vspace{-10pt}
    \caption{\textbf{Illustration of the DocHop data generation pipeline}. 
\textbf{(1) Metadata Pool}: We first prompt an LLM to curate a diverse pool of semantic metadata, including topics, entity names, candidate metrics, and timestamps that may appear in an instance. 
\textbf{(2) Chart Configuration Sampling}: From this pool, we sample a set of chart schemas by selecting chart types (e.g., bar, line) and instantiating their required structural fields from the metadata. 
\textbf{(3) Reasoning Trace Generation}: Conditioned on the sampled chart configurations, we stochastically construct a multi-hop reasoning trace, where each node corresponds to an instantiated constraint parameterized by a specific chart schema. 
\textbf{(4) Chart and Document Context Generation}: Given the symbolic trace, we prompt an LLM to generate constraint-satisfying chart tables and an accompanying narrative context that verbalizes the trace, while keeping the two processes independent. 
\textbf{(5) QA Curation and Document Rendering}: Finally, we assign a semantic reference label in the narrative, instantiate task-specific question--answer pairs grounded on this label, and render the charts and text into a single-page document image for evaluation.}
\label{fig:data_pipeline}
\end{figure*}

\section{DocHop}
\label{sec:dochop}
DocHop is constructed with a logic-first, reverse-engineered framework. The key design is to separate \emph{reasoning specification} from \emph{numerical evidence}: a symbolic reasoning trace defines the multi-step constraints, the document narrative verbalizes these constraints, and the charts provide the corresponding data values. To enforce joint chart-context reasoning, the narrative concludes by assigning a \emph{semantic reference label} to the entities implied by executing the trace under its logical composition, and all questions are grounded on this label rather than explicit entity names. This prevents direct chart-only lookup and requires models to resolve targets from the narrative before aggregating evidence from charts. We organize the construction process into four main components: (1) Symbolic Reasoning Trace Generation, (2) Chart and Document Context Generation, (3) Document Composition and QA Curation, and (4) Quality Verification.

\subsection{Symbolic Reasoning Trace Generation}

This subsection describes the first three stages of the pipeline: metadata pool curation, chart configuration sampling, and symbolic reasoning trace generation. The goal is to construct an abstract reasoning problem before any concrete chart values are generated.

We first establish a semantic context by prompting a large language model (LLM) to generate a diverse metadata pool. Each pool defines the global semantic space of an instance, including the document topic, candidate entities, measurable metrics, semantic groups that subdivide metric values, value units, and available timestamps. From this pool, we sample a set of $k$ chart schemas, where $k \in \{2, 3, 4, 6\}$. Each schema instantiates a chart structure by selecting metadata fields appropriate for a particular chart type, such as tracking a metric over time or comparing metric values across groups. At this stage, the schemas specify only the structural components of the charts, without any concrete numerical values; the values are generated later conditioned on the reasoning trace in the chart and document context generation stage.

Conditioned on the sampled chart schemas, we construct a symbolic reasoning trace modeled as a logical tree with a target reasoning depth $D \in \{2, 3, 4, 5\}$. The trace contains two types of nodes. Constraint nodes represent parameterized rules instantiated from templates and filter candidate entities based on chart-grounded conditions. Logical composition nodes combine entity sets from their input nodes using Boolean operations. Directed edges govern how candidate entity sets are passed through the tree. For a child constraint node, the node inherits the candidate entity pool from its parent and applies its rule to further filter this set. Consequently, the entities satisfying a child constraint node form a subset of those satisfying its parent. The trace is generated recursively through the following mechanisms:

\textbf{Node Parameterization.} To promote constraint diversity, we employ a library of 27 distinct rule templates (see Appendix~\ref{sec:appendix_rule_templates}) designed by human annotators. Each template is formulated as a parameterized function (e.g., \texttt{<Timestamp>}: \texttt{<Metric>} \texttt{<Operator>} \texttt{<Threshold>}) that must be instantiated with concrete values when sampled. During generation, the pipeline first samples a target subset of entities from the valid pool passed down by the parent node, and then populates the remaining slots (e.g., operator and threshold) to form a definite constraint (e.g., \textit{Products A and B satisfy the constraint ``2024-06: Items Sold $\ge$ 45''}). This ensures that the constraint is both structurally valid and logically consistent with the preceding context.

\textbf{Logical Composition.} Logical composition nodes aggregate constraints by combining entity sets from their input nodes. These nodes are parameterized by sampling a Boolean operator, specifically \textsc{AND} or \textsc{OR}. Unlike constraint nodes that filter entities through attribute-based conditions, the output entity set of a logical composition node is deterministic and logic-driven: it is computed by directly applying the sampled Boolean operator to its input entity sets.

The algorithm iteratively samples and composes these nodes until the target depth $D$ is reached. Due to the stochastic nature of this recursive generation, the process yields diverse reasoning skeletons, ranging from simple linear chains to complex multi-branch trees. The resulting logical structure specifies how entities should be filtered and combined across the sampled chart schemas, and serves as the blueprint for subsequent data synthesis. Examples of these diverse tree structures are provided in Appendix~\ref{sec:appendix_traces}.

\begin{figure*}[t]
    \centering
    \begin{subfigure}[b]{0.45\textwidth}
        \centering
        \includegraphics[width=\textwidth]{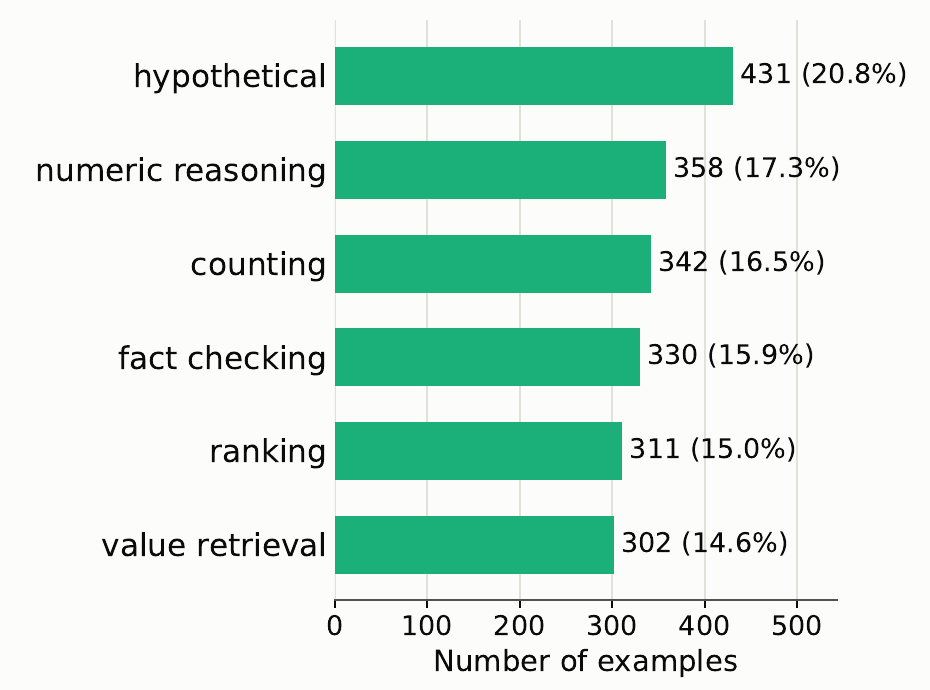}
        \caption{Distribution by Task}
        \label{fig:task_distribution}
    \end{subfigure}
    \hfill
    \begin{subfigure}[b]{0.45\textwidth}
        \centering
        \includegraphics[width=\textwidth]{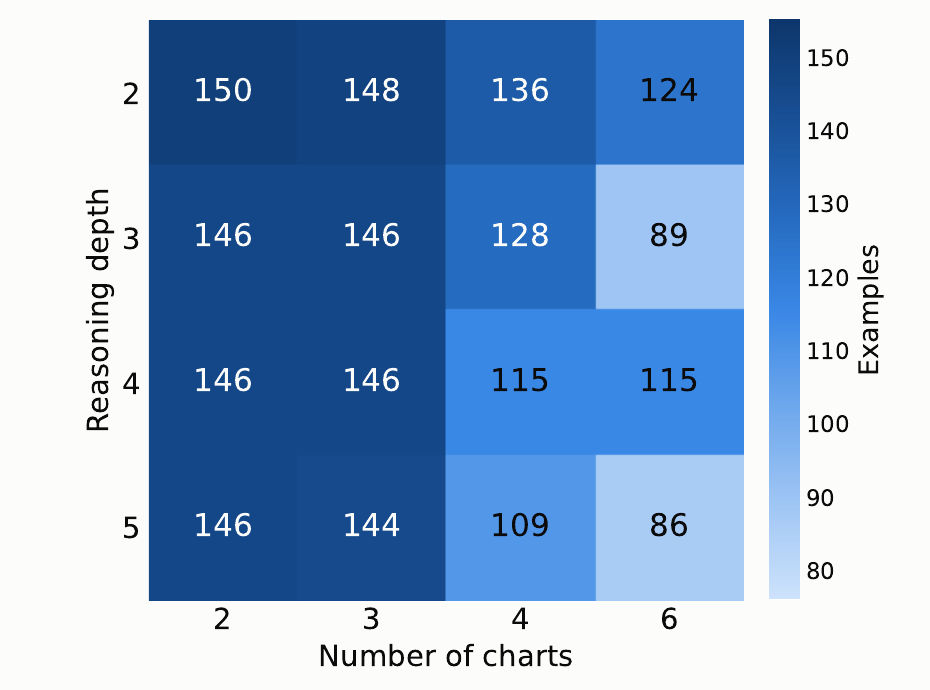}
        \caption{Distribution over Reasoning Depth and Number of Charts}
        \label{fig:distribution_depth}
    \end{subfigure}
    \caption{\textbf{Dataset distribution of \textsc{DocHop}}. We report the distribution across task categories (left) and the distribution over reasoning depths and numbers of charts per document (right).}
    \label{fig:dataset_stats}
\end{figure*}

\subsection{Chart and Document Context Generation}

Given a sampled reasoning trace and its associated chart schemas, we synthesize the two complementary sources of information in each \textsc{DocHop} instance: chart tables that provide numerical evidence, and document narratives that verbalize the reasoning specification. We generate these two components independently so that the document context describes the constraints without leaking the underlying chart values or the final entity assignments. We use Gemini-2.5-Pro for all LLM-based synthesis steps, and full prompt details are provided in Appendix~\ref{sec:prompts_data}.

\textbf{Chart Data Generation}. 
We generate numerical tables that satisfy the instantiated constraints in the reasoning trace. A naive approach would be to prompt the model to produce all charts jointly conditioned on the full multi-hop trace; however, as reasoning depth and chart count grow, it becomes difficult to reliably ensure that the generated tables satisfy all specified constraints simultaneously. To improve reliability, we adopt a per-chart synthesis strategy: we traverse the reasoning tree and aggregate all regular-node constraints associated with the same chart into a single requirement set. The prompt explicitly specifies which entities must satisfy or violate each condition, and the LLM generates the corresponding chart table independently for each chart. We then verify correctness by executing the reasoning trace programmatically. When converting these tables into chart images, we randomly sample a compatible chart type and color scheme for each chart, introducing diversification in both graphical form and appearance.

\textbf{Document Context Generation}. 
In parallel, we verbalize the symbolic reasoning trace into a coherent document narrative. Context generation is performed independently from chart synthesis: the LLM has no access to chart values or to the final entity set produced by executing the trace. Instead, it is provided only with the instantiated constraint structure, including the parameterized conditions at each node and their logical composition. This information is sufficient to describe the reasoning procedure in natural language, while preventing the narrative from revealing numerical evidence or directly naming the resulting target entities.

\subsection{Question and Answer Curation and Document Rendering}

\textbf{Semantic Reference Label}. 
After generating the document narrative, we assign a semantic reference label (e.g., \textit{Platinum Tier Night Service} in Figure~\ref{fig:teaser}) to the entities determined by executing the reasoning trace under its logical composition. Questions are grounded on this label rather than explicit entity names, preventing direct chart-only lookup and requiring models to first identify the relevant entity set from the narrative before aggregating numerical evidence from the charts. For instance, instead of asking for the average metric value of specific hotels (e.g., HTH, SCS, WG in Figure~\ref{fig:teaser}), a question is phrased as: ``What is the average \textit{Linens Replaced} under \textit{Night Shift} in \textit{August 2025} of the hotel(s) receiving the \textit{Platinum Tier Night Service Commendation}?''

\textbf{Task Definitions}. 
Following prior chart question answering benchmarks, we formulate \textsc{DocHop} questions into six reasoning categories:
\begin{itemize}\itemsep0pt
    \item \emph{Value Retrieval}: Retrieve a metric of entities assigned the semantic reference label, or directly ask which entities are assigned the semantic reference label.
    \item \emph{Counting}: Count how many entities are assigned the semantic reference label, optionally under an additional chart-grounded condition specified in the question.
    \item \emph{Numeric Reasoning}: Perform arithmetic aggregation (e.g., sum or average) over chart entries of entities assigned the semantic reference label.
    \item \emph{Ranking}: Identify extrema or order entities assigned the semantic reference label based on their chart values.
    \item \emph{Hypothetical Reasoning}: Answer counterfactual queries that modify either chart values while keeping the semantic reference label fixed, or the reasoning trace itself by introducing additional constraints.
    \item \emph{Fact Checking}: Verify whether a statement is entailed by jointly applying the textual constraints and chart evidence over entities assigned the semantic reference label.
\end{itemize}

\textbf{Question and Answer Construction}. 
For each task type, we start from a corresponding question template and instantiate it using the synthesized document metadata. These templates (41 in total) are designed by human annotators to cover diverse reasoning patterns across tasks. Questions are grounded on the semantic reference label defined in the narrative rather than explicit entity names in the charts, ensuring that answering requires integrating information from the narrative context and the associated charts. For example, a numeric reasoning question may ask: ``What is the total \textit{Transaction Costs Paid} in \textit{December 2024} of all investment funds receiving the \textit{Certified Prudent Operator designation}?''. Since the reasoning trace deterministically defines the target entity set, ground-truth answers are computed exactly by executing the corresponding operations over the generated chart tables. We provide the full list of question templates under each task in Appendix~\ref{sec:appendix_question_templates}.

\textbf{Document Rendering}. 
We render the synthesized document narrative and chart images into a single-page document using ReportLab with a fixed A4-style layout; the question and answer are kept separately and are not included in the rendered document. To ensure that each instance fits on one page, we start from font size 10 and progressively decrease the font size until the narrative fits. We also adjust chart scaling across different subplot layouts so that individual subplots remain roughly comparable in visual size. The final document is rendered at 300 DPI and paired with its corresponding question and verified answer to form one \textsc{DocHop} evaluation instance. The entire data curation pipeline is illustrated in Figure~\ref{fig:data_pipeline}.

\subsection{Quality Verification}

We apply a multi-stage verification process to ensure the correctness and readability of \textsc{DocHop} instances.

\textbf{Programmatic Verification of Chart Data}. 
Since the symbolic reasoning trace and generated chart tables are fully accessible, we verify each instance by executing the trace against the chart data. This allows us to deterministically check whether the generated numerical values satisfy all instantiated constraints and whether the resulting target entity set and ground-truth answer are correct. Instances that fail this verification are discarded and regenerated.

\textbf{Human Verification of Document Context}. 
We then manually inspect whether the generated document narrative faithfully verbalizes the underlying reasoning trace. This step checks whether the textual constraints, logical compositions, and semantic reference label are expressed clearly and consistently with the symbolic trace. Instances with missing, ambiguous, or incorrect descriptions of the reasoning process are rejected.

\textbf{Final Rendering Verification}. 
Finally, we inspect the rendered document images to ensure that the document context and charts are fully visible and readable. This includes checking for truncated text, missing context, overlapping elements, occluded charts, illegible labels, or other layout artifacts introduced during PDF rendering. Instances that fail the final rendering check are regenerated.

\subsection{Dataset Statistics}
DocHop contains 2,074 document instances, each paired with a single question-answer example. We construct the benchmark with controllable reasoning complexity along two axes: reasoning depth $D \in \{2,3,4,5\}$ and the number of charts per document $k \in \{2,3,4,6\}$. Each example is assigned to one of six task categories (Value Retrieval, Counting, Numeric Reasoning, Ranking, Hypothetical Reasoning, and Fact Checking), instantiated from a library of question templates.

During generation, we aim to balance the sampling across task types as well as across different depth and chart-count configurations, enabling systematic evaluation over a broad range of reasoning difficulty. The example distributions of DocHop can be found in Figure~\ref{fig:dataset_stats}.

\begin{table*}[ht!]
\centering
\small
\setlength{\tabcolsep}{5pt}
\resizebox{\textwidth}{!}{
\begin{tabular}{lccccccc}
\toprule
\textbf{Model} & \textbf{Value Retrieval} & \textbf{Counting} & \textbf{Numeric Reasoning} & \textbf{Ranking} & \textbf{Hypothetical} & \textbf{Fact Checking} & \textbf{Overall} \\
\midrule
Human Eval & 95.36 & 94.15 & 95.53 & 93.56 & 86.54 &96.67& 93.29 \\
Random Guessing (Gemini~\cite{Gemini2.5}) & 0.00 & 4.97 & 0.69 & 0.00 & 9.21 & 45.56 & 10.10 \\
Random Guessing (GPT~\cite{openai2025gpt52}) & 0.00 & 8.35 & 0.23 & 0.00 & 10.79 & 47.83 & 11.27 \\
\midrule
\multicolumn{8}{l}{\textbf{Proprietary Reasoning Models}} \\
GPT-5.2-Reasoning~\cite{openai2025gpt52} & \textbf{66.56} & \textbf{67.84} & \textbf{46.93} & \textbf{60.77} & \textbf{58.70} & \textbf{78.79} & \textbf{62.83} \\
GPT-5-mini-Reasoning~\cite{GPT5} & 29.14 & 23.68 & 9.50 & 31.19 & 23.67 & 55.76 & 28.25 \\
Gemini-2.5-Pro-Reasoning~\cite{Gemini2.5} & 41.39 & 49.12 & 17.60 & 39.55 & 35.73 & 63.33 & 40.60 \\
Gemini-2.5-Flash-Reasoning~\cite{Gemini2.5} & 35.10 & 40.94 & 11.17 & 28.62 & 22.27 & 58.48 & 32.02 \\
\midrule
\multicolumn{8}{l}{\textbf{Proprietary Models}} \\
GPT-5.2~\cite{openai2025gpt52} & 40.73 & 56.14 & 19.83 & 37.30 & 35.50 & 55.15 & 40.36 \\
Gemini-2.5-Flash~\cite{Gemini2.5} & 27.15 & 25.73 & 8.94 & 28.30 & 16.94 & 46.36 & 24.88 \\
Claude-4.5-Haiku~\cite{anthropic2025claudehaiku45} & 2.98 & 19.01 & 1.96 & 4.82 & 11.37 & 58.79 & 16.35 \\
Claude-4.5-Sonnet~\cite{anthropic2025claudesonnet45} & 4.64 & 28.65 & 2.51 & 7.72 & 10.44 & 55.15 & 17.94 \\
\midrule
\multicolumn{8}{l}{\textbf{Open-Source Models}} \\
LLaVA-Next-LLaMA3-8B~\cite{liu2024llavanext} & 0.00 & 9.65 & 0.56 & 0.00 & 8.82 & 49.09 & 11.33 \\
IDEFICS2-8B~\cite{laurenccon2024matters} & 2.32 & 4.39 & 1.40 & 1.61 & 8.12 & 35.45 & 8.87 \\
IDEFICS3-LLaMA3-8B~\cite{laurençon2024building} & 7.62 & 9.36 & 1.96 & 3.86 & 5.57 & 37.27 & 10.66 \\
Qwen-2.5-VL-7B~\cite{bai2025qwen2} & 16.89 & 13.74 & 4.19 & 24.12 & 12.06 & 46.97 & 19.05 \\
Qwen3-VL-8B~\cite{Qwen3-VL} & 16.89 & 22.51 & 10.61 & 32.15 & 16.47 & 49.70 & 24.16 \\
Molmo-7B-O-0924~\cite{deitke2025molmo} & 4.30 & 23.98 & 2.51 & 12.22 & 11.60 & 46.97 & 16.73 \\
Molmo-7B-D-0924~\cite{deitke2025molmo} & 9.93 & 21.05 & 2.23 & 13.18 & 10.21 & 41.52 & 16.01 \\
Ovis1.6-Gemma2-9B~\cite{lu2024ovis} & 1.32 & 8.48 & 0.28 & 1.93 & 7.89 & 43.94 & 10.56 \\
InternVL-3.5-8B~\cite{internvl35} & 3.31 & 10.23 & 1.96 & 4.82 & 8.58 & 53.03 & 13.45 \\
InternVL-3.5-30B-A3B~\cite{internvl35} & 2.32 & 19.30 & 2.79 & 2.89 & 11.83 & 49.39 & 14.75 \\
Qwen-2.5-VL-32B~\cite{bai2025qwen2} & 21.85 & 19.01 & 6.70 & 22.51 & 16.47 & 47.27 & 21.79 \\
InternVL-3.5-38B~\cite{internvl35} & 4.97 & 21.93 & 4.19 & 3.86 & 9.05 & 50.61 & 15.57 \\
Qwen-2.5-VL-72B~\cite{bai2025qwen2} & 23.84 & 26.32 & 7.82 & 25.08 & 17.87 & 48.79 & 24.40 \\
\bottomrule
\end{tabular}
}
\caption{\textbf{Evaluation results on DocHop across task categories}. We report both per-task accuracy and the overall accuracy of DocHop.}
\label{tab:dochop_results}
\end{table*}

\subsection{Evaluation Metrics}

We evaluate model performance using normalized answer accuracy. For non-numeric questions, we use case-insensitive exact match. For numeric questions, we normalize outputs to a canonical format and apply tolerance-based matching under the required precision. For binary fact-checking questions, accuracy is computed over Yes/No predictions. Additional evaluation details, such as prompting instructions, answer parsing rules, and numeric precision handling, are provided in the Appendix~\ref{sec:appendix_eval}.

\section{Experiments}

\subsection{Evaluation Setup}

\begin{figure*}[ht]
    \centering
    \includegraphics[width=\textwidth]{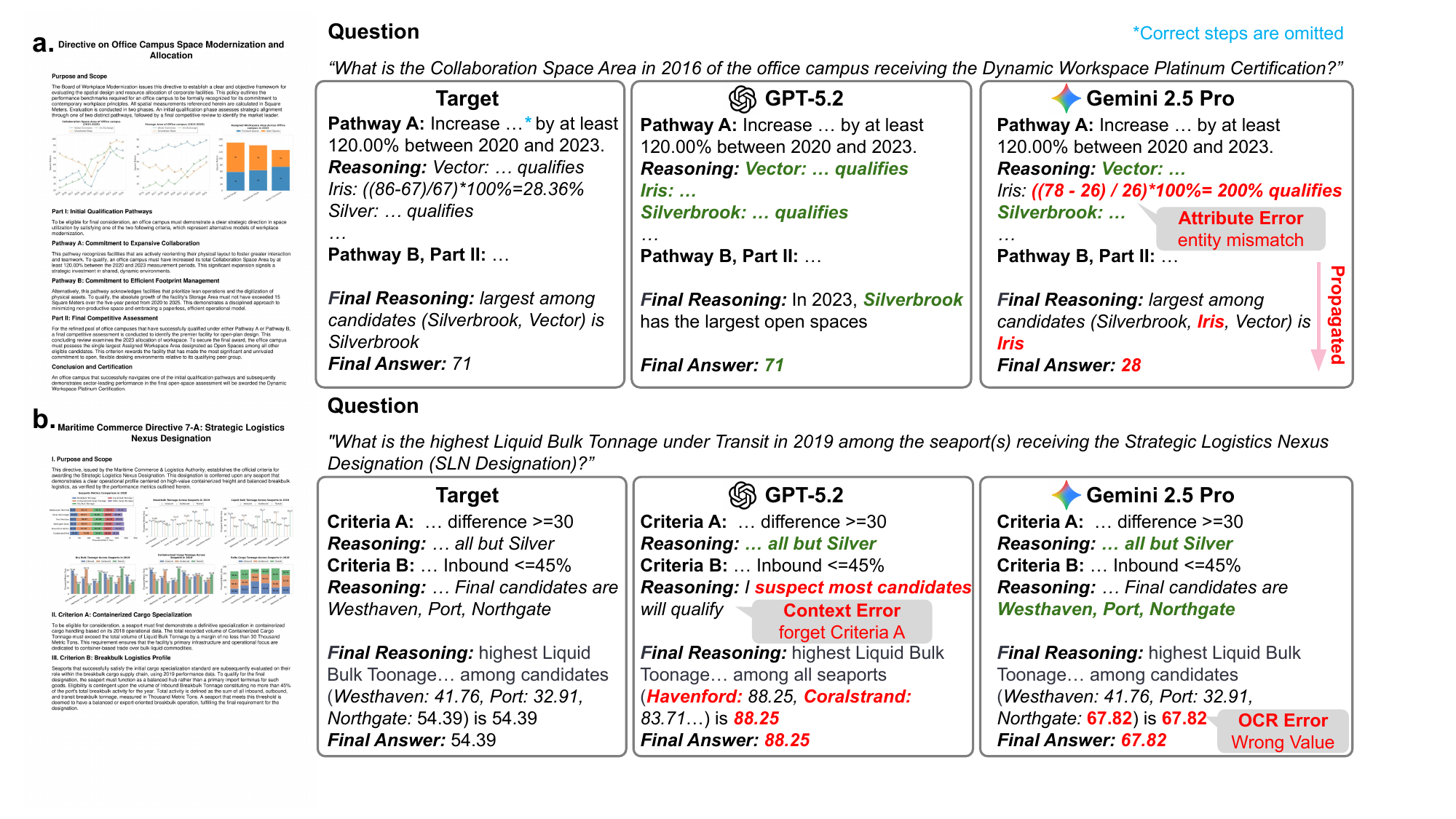}
    \vspace{-13mm}
    \caption{\textbf{Qualitative examples of reasoning outputs}. 
(a) An instance where GPT-5.2 Reasoning answers correctly, while Gemini-2.5-Pro fails by attributing chart values to the wrong entity, leading to an incorrect aggregation. 
(b) A challenging case where both models make incorrect predictions but for different reasons: GPT identifies the global maximum instead of restricting to entities receiving the semantic label, whereas Gemini resolves the correct entity set from the narrative but makes an OCR mistake in the final value extraction.}
    \label{fig:qualitative}
\end{figure*}

\textbf{Evaluated Models}. 
We evaluate a broad range of state-of-the-art MLLMs, including both proprietary and open-source systems. Our proprietary model set includes GPT-5.2~\cite{openai2025gpt52}, GPT-5-mini~\cite{GPT5}, Gemini-2.5-Pro/Flash~\cite{Gemini2.5}, Claude 4.5 Haiku~\cite{anthropic2025claudehaiku45}, and Claude 4.5 Sonnet~\cite{anthropic2025claudesonnet45}. We additionally report results for their reasoning-enhanced variants when available (e.g., GPT-5.2-Reasoning, Gemini-2.5-Pro-Reasoning, Gemini-2.5-Flash-Reasoning). 

For open-source baselines, we consider representative vision--language models across multiple families and scales, including LLaVA-Next-LLaMA3-8B~\cite{liu2024llavanext}, IDEFICS2-8B~\cite{laurenccon2024matters}, IDEFICS3-LLaMA3-8B~\cite{laurençon2024building}, Qwen-2.5-VL (7B/32B/72B)~\cite{bai2025qwen2}, Qwen3-VL-8B~\cite{Qwen3-VL}, Molmo-7B (O/D)~\cite{deitke2025molmo}, Ovis1.6-Gemma2-9B~\cite{lu2024ovis} and InternVL-3.5 (8B/30B-A3B/38B)~\cite{internvl35}. Full model configuration settings are provided in Appendix~\ref{sec:model_config}.

\textbf{Human and Random Guessing Baselines}. 
We obtain human performance by hiring graduate student annotators, who are presented with the full document image and asked to provide free-form answers. Following ChartXiv~\cite{charxiv}, we also obtain a \emph{question-only} guessing baseline by prompting GPT-5.2~\cite{openai2025gpt52} and Gemini-2.5-Pro~\cite{Gemini2.5} with only the question text, without providing the document image. This serves as a language-prior baseline.

\subsection{Main Results}
Table~\ref{tab:dochop_results} summarizes the full evaluation results. Here, we summarize the main takeaways.

\textbf{Overall Gap to Human Performance}. 
All evaluated models, including both proprietary and open-source ones as well as reasoning variants, remain far below human accuracy. While human annotators achieve over 90\% accuracy, the best-performing model (GPT-5.2 Reasoning) reaches only 62.83\% overall, indicating a substantial gap in integrated chart-context reasoning. It is also worth noting that it takes approximately five minutes per question on average for human annotators to solve one question, suggesting that DocHop requires careful multi-step reasoning rather than quick visual lookup. The large discrepancy between human and model performance highlights the difficulty of document-level multi-hop reasoning for current MLLMs.

\textbf{Reasoning Models Consistently Outperform Their Non-Reasoning Counterparts}. 
As shown in Table~\ref{tab:dochop_results}, models with explicit reasoning mechanisms achieve substantial gains over their base variants across both GPT and Gemini families. For example, GPT-5.2 improves from 40.36\% to 62.83\% overall when switching to its reasoning variant, and Gemini-2.5-Flash increases from 24.88\% to 32.02\%. Gemini-2.5-Pro with a dynamic reasoning mechanism further reaches 40.60\% overall accuracy, outperforming the Gemini-2.5-Flash variants. Overall, these results indicate that explicit reasoning remains an important factor for stronger performance on DocHop.

\textbf{Proprietary Models Remain Stronger Overall}. 
Among all evaluated models, GPT-5.2 Reasoning achieves the best overall performance, reaching 62.83\% accuracy on DocHop. Gemini-2.5-Pro Reasoning is the next strongest proprietary baseline at 40.60\%, though a substantial gap remains compared to GPT. More broadly, proprietary models generally outperform the open-source baselines in our current evaluation. Most open-source models achieve overall accuracies in the range of roughly 9--24\%, indicating that integrated chart--context reasoning remains challenging for current open vision--language models.

\subsection{Analysis}

\begin{figure*}[t]
    \centering
    \begin{subfigure}[b]{0.4\textwidth}
        \centering
        \includegraphics[width=\textwidth]{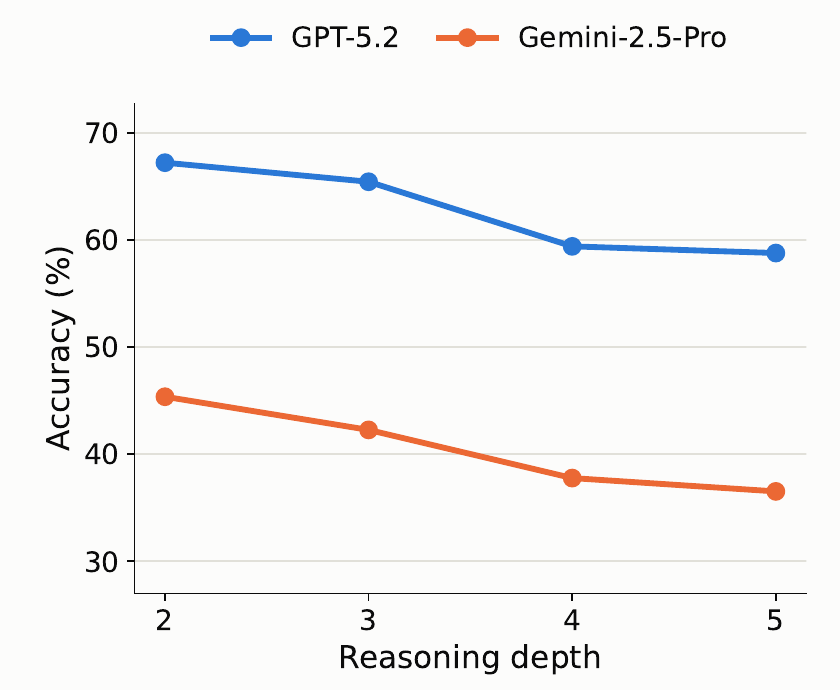}
        \caption{Performance by Reasoning Depth}
        \label{fig:reasoning_depth}
    \end{subfigure}
    \hfill
    \begin{subfigure}[b]{0.4\textwidth}
        \centering
        \includegraphics[width=\textwidth]{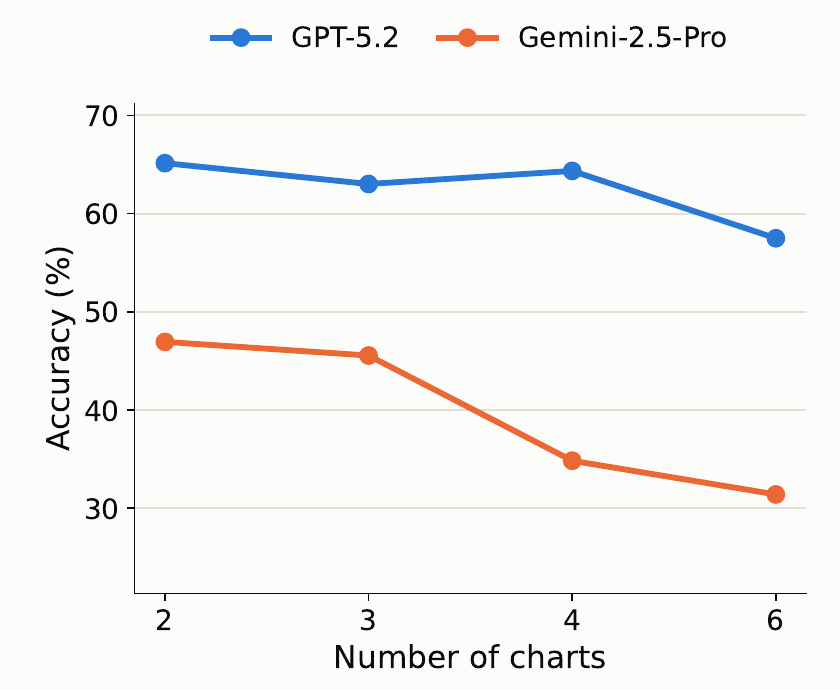}
        \caption{Performance by Number of Charts}
        \label{fig:chart_num}
    \end{subfigure}

    \caption{\textbf{Performance of GPT-5.2 and Gemini-2.5-Pro under controlled DocHop complexity}.
    We report accuracy of GPT-5.2 Reasoning~\cite{openai2025gpt52} and Gemini-2.5-Pro Reasoning~\cite{Gemini2.5}, both operating with explicit reasoning enabled, as reasoning depth and number of charts increase.
    Performance generally degrades for both models as either the reasoning trace becomes deeper or the document becomes visually denser, highlighting the increasing difficulty of cross-chart evidence aggregation under narrative constraints.}
    \label{fig:reasoning_performance}
\end{figure*}

\textbf{Model performance degrades with more charts and higher reasoning depth}.
Figure~\ref{fig:reasoning_performance} shows that the performance of both GPT-5.2 Reasoning and Gemini-2.5-Pro Reasoning consistently decreases as the number of charts in the document and the reasoning depth increase. Accuracy is highest in simpler settings with fewer charts ($k=2$) and shallow traces ($D=2$), and drops steadily as the document becomes visually denser and the reasoning trace requires more hops. This trend suggests that scaling DocHop along either axis imposes additional difficulty for current MLLMs in aggregating evidence across multiple charts under narrative constraints.

\textbf{Qualitative Examples.}
Figure~\ref{fig:qualitative} shows two representative DocHop instances. \textbf{(a) Attribution Error.} In Figure~\ref{fig:qualitative}(a), GPT-5.2 Reasoning model answers correctly by grounding the reasoning steps on the right entity throughout. Gemini-2.5-Pro, however, makes an attribution mistake: it confuses the chart values of two entities, leading to an incorrect aggregation despite otherwise plausible intermediate reasoning. \textbf{(b) Different Failure Modes on the Same Instance.}
Figure~\ref{fig:qualitative}(b) illustrates a case where both models fail, but for different reasons. GPT-5.2 Reasoning loses track of the narrative constraint and reports the global maximum rather than the maximum among entities receiving the SLN Designation. Gemini-2.5-Pro correctly identifies the qualified entity set, but produces an incorrect final answer due to a misread chart value. These examples suggest that DocHop remains challenging even for strong proprietary models, with errors in both contextual reasoning and chart understanding.

\section*{Limitations and Future Work}

\textsc{DocHop} is designed as a controlled out-of-domain reasoning benchmark rather than a dataset that reproduces the distribution of naturally occurring documents. This design allows us to isolate chart--document integration under controllable reasoning depth and visual density, but it also means that performance on \textsc{DocHop} should be interpreted as a diagnostic measure of a specific reasoning capability rather than as a direct estimate of performance on real-world document collections. Future work can extend this logic-grounded evaluation framework to naturally occurring reports, scientific articles, business dashboards, and other document domains.

In addition, \textsc{DocHop} focuses on chart-based numerical evidence paired with narrative context. Real documents often contain other structured visual elements, such as tables, diagrams, forms, equations, and user-interface screenshots. Extending the logic-grounded construction framework to these document elements would enable broader evaluation over heterogeneous document structures. Our current rendering pipeline also uses controlled single-page A4-style layouts with readable charts and narrative text. This design reduces confounding factors from severe OCR noise, document parsing failures, and layout artifacts, allowing the benchmark to focus on chart--document reasoning; however, it does not capture the full visual variability of naturally occurring documents. Future work can introduce multi-page documents, noisier scans, and more diverse layout styles. Finally, although our construction pipeline combines programmatic verification with human review, synthetic benchmark generation may still leave occasional annotation, wording, or rendering issues. We plan to maintain the released benchmark with versioned updates and incorporate validated community feedback in future releases.

\section{Conclusion}

We introduced \textsc{DocHop}, a benchmark for integrated chart--context reasoning in document-style images, where models must resolve narrative constraints and aggregate numerical evidence across multiple charts. 
A key design is the use of semantic reference labels grounded in the document narrative, which prevents direct chart-only lookup and enforces joint reasoning over text and visual data. 
Experiments across both proprietary and open-source MLLMs reveal a substantial gap to human performance, while reasoning-enhanced variants consistently improve accuracy over their base counterparts. 
We further observe a steady degradation as reasoning depth and chart count increase, highlighting persistent challenges in multi-hop evidence aggregation under dense visual contexts. 
Future work can extend this logic-grounded evaluation to other structured visual elements (e.g., tables, diagrams, and UI documents) to better characterize multimodal reasoning in realistic settings.

\section*{Impact Statement}

This paper presents work whose goal is to advance the field of multimodal machine learning through the development of \textsc{DocHop}, a benchmark for evaluating integrated chart--context reasoning in document images. 
Our work is primarily intended to support more reliable assessment of multimodal reasoning capabilities, and does not introduce new deployment-facing model components. 
Potential societal impacts are therefore indirect, such as improving the robustness and transparency of evaluation for systems used in document understanding. 
As with other synthetic benchmark datasets, misuse could arise if models are over-optimized for benchmark performance without corresponding real-world generalization. 
We encourage future work to consider broader coverage of document domains and to complement benchmark-driven progress with responsible evaluation practices.

\section*{Acknowledgment}
This work was supported in part by NSF IIS2404180, IBM, Institute of Information \& communications Technology Planning\& Evaluation (IITP) grant funded by the Korea government (MSIT) (No. 2022-0-00871, Development of AI Autonomy and Knowledge Enhancement for AI Agent Collaboration), and Electronics and Telecommunications Research Institute (ETRI) grant (26CB1200, Development and Application of Science-Specialized Multimodal Foundation Models).

\nocite{langley00}

\bibliography{example_paper}
\bibliographystyle{icml2026}

\newpage
\onecolumn
\appendix
\counterwithin{table}{section}
\counterwithin{figure}{section}

\section{LLM Prompts for Data Curation}
\label{sec:prompts_data}

We provide the prompt templates used in the construction of DocHop. 
Given a sampled symbolic reasoning trace (Section~\ref{sec:dochop}), we employ an LLM to synthesize both the underlying chart tables and the accompanying document narrative. To ensure reliability and prevent information leakage, chart generation and context generation are performed independently: chart prompts include explicit constraint satisfaction requirements, while narrative prompts only receive the instantiated constraint structure without access to chart values or entity assignments.

Across the pipeline, we use three main classes of prompts:
(i) \textbf{Metadata Pool Curation}, which generates diverse topics, entities, metrics, and timestamps;
(ii) \textbf{Per-chart Table Synthesis}, which produces numerical tables satisfying aggregated chart-specific constraints;
and (iii) \textbf{Narrative Context Generation}, which verbalizes the reasoning trace into policy-style document text and introduces the semantic reference label used for question grounding.
All prompts are executed with Gemini-2.5-Pro.

\lstset{
    basicstyle=\ttfamily\small,
    breaklines=true,
    breakatwhitespace=false,
    columns=flexible,
    keepspaces=true,
    showstringspaces=false,
    frame=single,
    backgroundcolor=\color{gray!10}
}

\vspace{0.5em}
\noindent\textbf{Metadata Pool Prompt.}

\begin{lstlisting}
You are MetaSetWriter. Return ONLY a single JSON object.

TOPIC: "{topic}"

WHAT TO RETURN
{
  "entity_type": "<singular type label>",
  "entities": ["<E names>"],
  "groups":   ["<G orthogonal subgroups>"],
  "metrics":  ["<M independent metrics in SAME family>"],
  "unit":     "<A concrete plural noun representing the measurement unit>",
  "data_type": "<'integer' OR 'float'>",
  "timestamps": ["<T strictly increasing stamps (e.g., Years '2020' or Months '2022-01')>"]
}

LOGIC RULES (CRITICAL)

1. NO MATHEMATICAL REDUNDANCY (INDEPENDENCE):
   - Metrics must be independent variables so that random values do not contradict each other.
   - THE "PART vs. WHOLE" RULE:
     - CONFLICT: Do NOT include both a "Total" and its "Parts" in the same list.
       - Bad: ["Total Revenue", "Service Revenue", "Product Revenue"] 
              (Because Product + Service might not equal Total in random data).
     - Solution A (Components Only): ["Service Revenue", "Product Revenue", "Consulting Revenue"].
     - Solution B (Aggregates Only): ["Total Revenue", "Total Cost", "Total Tax"].
   - THE "FORMULA" RULE:
     - CONFLICT: Do NOT include variables that are simple math functions of others.
       - Bad: ["Revenue", "Cost", "Profit"] (Because Profit = Revenue - Cost).
     - Fix: Pick a level of abstraction. E.g., just ["Revenue", "Cost", "Marketing Spend"] 
            OR just ["Net Profit", "EBITDA"].

2. DESCRIPTIVE METRIC NAMES (NATURAL LANGUAGE FRIENDLY):
   - Choose metric names that sound natural in a sentence (e.g., "The <Metric> was high").
   - GOOD: "Bottles Produced", "Patients Admitted", "Parcels Delivered".
   - BAD: "Total Produced", "Count", "Number".
   - Note: It is OK if the Metric name repeats the Unit name 
           (e.g., Metric: "Bottles Produced", Unit: "Bottles") if it improves clarity.

3. UNIT SCALING & SPECIFICITY:
   - NO generic "counts" or "values". Use specific nouns: "Parcels", "Patients", "Tons", "Joules".
   - SCALING: Use scale prefixes (Thousand, Million, Billion) for ANY unit if the entity scale justifies it.
     - "Million USD" (for Corporations)
     - "Thousand Liters" (for Water Plants)
     - "Metric Tons" (for Mining)
     - "Million Users" (for Tech Platforms)
   - NO RATES: NO "percent" or "ratio". Metrics must be summable.

4. GROUPS MUST BE UNIVERSAL CATEGORIES (ORTHOGONAL):
   - THE GOLDEN RULE: Every single Metric must make sense for Every single Group.
   - GOOD: Groups=["North", "South"] (Applies to almost anything).
   - BAD: Metric="Virus Cases", Groups=["Viral", "Bacterial"] 
          (A Virus cannot be Bacterial -> Logic Fail).
   - Groups must act as buckets that categorize the metrics, regardless of what the metric is.


SAFETY & CONTENT GUIDELINES (STRICT)

1. NO BIOLOGICAL FLUIDS/GORE:
   - Ban: "Blood", "Organs", "Bodily Fluids", "Kills".
   - Alt: "Essence", "Samples", "Elixir", "Units".

2. NO VIOLENCE/HARM:
   - Ban: "Fatalities", "Casualties", "Deaths".
   - Alt: "Incidents", "Transfers", "Losses", "Expirations".

3. NO SENSITIVE REAL-WORLD TOPICS:
   - Avoid real religious figures, political hate speech, or illicit contraband.
   - Keep fantasy/cult topics PG-13 (e.g., "Mana", "Relics").

4. NO REAL-WORLD ENTITIES (USE FICTIONAL REALISM):
   - Ban: Real country names (e.g., USA, China), real cities (e.g., Paris, Tokyo), 
          real famous people/politicians, or real company names.
   - Ban: Lazy placeholders (e.g., "Region A", "Person X", "Country 1").
   - Do: Invent realistic-sounding but fictional names 
         (e.g., "Westhaven", "Port Meridian", "Director Vance", "Silver Creek", "The Nordic Union").

{entity_type_clause}

IN-CONTEXT EXAMPLE A (Financial/Annual - No Math Conflicts)
{
  "entity_type": "conglomerate",
  "entities": ["Aether Corp", "Nebula Logistics", "Terra Firm"],
  "groups": ["North America", "EMEA", "APAC"],
  "metrics": ["Global Revenue", "Liquidity Assets", "Market Cap"],
  "unit": "Million USD",
  "data_type": "float",
  "timestamps": ["2018","2019","2020","2021","2022","2023"]
}

IN-CONTEXT EXAMPLE B (Operational/Monthly - Scaled Physical Unit)
{
  "entity_type": "bottling_plant",
  "entities": ["Vortex Springs", "Alpine Pure", "Crystal Cove"],
  "groups": ["Glass Bottles", "PET Plastic", "Aluminum Cans"],
  "metrics": ["Bottles Produced", "Bottles Scrapped", "Bottles Shipped"],
  "unit": "Thousand Bottles",
  "data_type": "integer",
  "timestamps": ["2023-01","2023-02","2023-03","2023-04","2023-05","2023-06"]
}
\end{lstlisting}

\vspace{0.5em}
\noindent\textbf{Data Synthesis Prompt--Line Plot as Example.}

\begin{lstlisting}
You must generate the chart below. The chart must strictly follow the chart type, metrics, timestamps, groups, and rules assigned to it. All entity-level constraints must be satisfied exactly.

IMPORTANT CONSTRAINT INTERPRETATION:
- When a rule specifies that certain entities MUST satisfy a constraint, those entities must strictly satisfy it.
- Entities NOT mentioned in a constraint MUST VIOLATE that constraint.
- For example: 'For entities A, B: value >= 70' means A and B must have value >= 70, while all other entities must have value < 70.

You must return the final answer in this JSON format:

{
  "{chart_id}": "<csv string>"
}

Where the <csv string> is valid CSV, with:
- First column = entity name
- **All numerical values are in: {unit}**.
- Note the scale '{unit}': A value of 50 means 50 {unit}.
- Column headers MUST be exactly the metric/group names provided. DO NOT append the year (e.g. use 'Revenue', NOT 'Revenue (2017)').
- No commentary, no markdown, no chain-of-thought, no explanation.
- You MUST NOT leave the chart empty.
- CSV string may be long; that is allowed.
- You MUST output real values for the chart.
- Each value should be a float rounded to two decimal places between 20.00 and 99.99.
- Do not make all values end with .00 or .50 to avoid looking artificial.
- Example endings can include .23, .87, .14, .69, etc.

Bad examples (DO NOT output):
20.00, 21.00, 22.00, 23.50, 24.50

Good examples (follow these patterns):
20.13, 21.47, 22.84, 23.26, 24.79

CRITICAL VISUALIZATION REQUIREMENT FOR LINE CHARTS:
- When plotted, the lines must be visually distinguishable.
- Try your best to curate values so that at each timestamp, no two entities will have close values.
- i.e, try to avoid two points being too close when plotting the data.

Example Output 1:
{
  "{chart_id}": "Entity,2017,2018,2019,2020,2021,2022,2023,2024,2025\n
A,85.23,87.91,86.45,88.17,87.29,89.73,88.42,86.84,87.56\n
B,52.84,58.39,49.17,55.73,61.29,47.84,54.18,59.62,51.47\n
C,38.47,42.91,46.23,50.18,53.84,57.29,60.73,64.18,67.92\n
D,23.91,26.47,29.18,31.84,28.73,32.19,35.42,33.76,36.84"
}

NOTE: A (high 80s, stable), B (mid 50s, highly volatile),
      C (rising from 38 to 68), D (low 30s, slow rise)

Example Output 2:
{
  "{chart_id}": "Entity,2017,2018,2019,2020,2021,2022,2023,2024,2025\n
A,89.23,84.67,79.41,74.85,70.29,66.73,63.18,59.84,56.47\n
B,44.81,41.29,38.73,36.92,39.47,43.18,47.84,52.29,56.73\n
C,67.34,71.92,76.18,79.84,75.29,70.47,65.91,61.28,57.73\n
D,28.47,24.91,21.73,25.29,29.84,34.18,38.73,43.29,47.91"
}

NOTE: A (high 70s, steady decline), B (mid 40s, U-shaped),
      C (high 60s-70s, inverted-U), D (low 30s, V-shaped)

Both examples show diverse patterns - your data should also have varied trends.

---

## CHART {chart_id}
Type: LINE
Entities: {entity1, entity2, entity3, ...}
Unit: {unit}
Metric: {metric_name}
Timestamps: {2017, 2018, 2019, 2020, 2021, 2022, 2023, 2024, 2025}

Rules:
{constraint_1}
{constraint_2}
{constraint_3}
...
\end{lstlisting}

\noindent\textbf{Context Curation Prompt.}

\begin{lstlisting}
You are an expert Policy Drafter. Your goal is to translate a decision tree (provided in DOT format) into a professional, natural language policy document.

Return a single JSON object:
{
  "title": "A professional, bureaucratic title for the policy",
  "final_label": "The specific name of the final status/award",
  "text": "The full policy text in Markdown format..."
}

I. Generic Graph Grammar & Narrative Rules

1. Serial Topology (The Filtering Pipeline)
   * Structure: `Node A -> Node B` (Direct arrow connection).
   * Interpretation: A strict dependency. Node B is not an independent step; it operates 
     only on the subset of entities that have successfully passed Node A.
   * Narrative Requirement: You must explicitly signal Inherited Eligibility at the start 
     of the second node's description. Clarify that the subsequent criteria apply 
     exclusively to the specific pool of candidates that survived the previous filter.

2. Convergent Topology (Combinatorial Logic)
   * Structure: Multiple arrows converging into a single `COMBO` node.
   * Edge Order Rule: The order of incoming edges in the `dot_string` defines the 
     operand roles (Crucial for Subtraction):
     - 1st Edge: Left Operand (Input A).
     - 2nd Edge: Right Operand (Input B).

3. Logic Definitions & Narrative Requirements
   * AND (Intersection)
     - Meaning: Simultaneous satisfaction of all inputs.
     - Req: Describe as a unified, mandatory standard where no criterion is optional.

   * OR (Union)
     - Meaning: Any input suffices.
     - Req: Describe as alternative pathways or flexible qualification routes.

II. NARRATIVE ARCHITECTURE (Randomly assigned)
You must follow the specific structure directed in the user prompt:
1. STRUCTURE_A (Conclusion-Led): State the {final_label} in the first paragraph as 
   the policy's primary goal. Then, detail the prerequisite requirements.
2. STRUCTURE_B (Criteria-Led): Detail the requirements phase-by-phase, introducing 
   the {final_label} only in the concluding summary.

III. WRITING GUIDELINES (FORMULA-TO-POLICY TRANSLATION)
- STORYTELLER, NOT CALCULATOR: Do not simply list data points. Write the policy as 
  a cohesive narrative about the {entity_type}'s performance history and compliance 
  expectations.
- FORMULA IS TRUTH: The user provides logic nodes in mathematical notation 
  (e.g., A/B >= 50%). These are the absolute ground truth. Your job is to translate 
  these mathematical relationships into professional policy prose.
- DE-MECHANIZATION: Do not simply read the formula aloud.
  * BAD: "The value of X divided by the sum of Y must be greater than 0.5."
  * BAD: If a node's formulate requires comparing with other entities, DO NOT WRITE: 
    "calculated across all entities in the dataset". USE "all participating 
    {entity_type}" instead.
- PROFESSIONAL PROSE: You have full creative freedom to rephrase the technical labels 
  from the DOT nodes into smooth, authoritative bureaucratic language.
- TERMINOLOGY: Replace generic "entity" with the specific `entity_type` provided 
  (e.g., "contract", "farm").
- HANDLING EXTREME VALUES (MAX/MIN Constraints):
  * Context: If a node rule says "value == the max value" or "value == the min value".
  * Requirement A (Competitive Ranking): You MUST interpret this as a ranking. 
    Explicitly state that the entity recorded the highest (or lowest) absolute figure 
    in the comparative set. Use keywords like "Unrivaled," "Market-leading," 
    "Highest recorded value."
  * Requirement B (Scope Abstraction - CRITICAL): If the rule lists specific entity 
    names (e.g., "among Entity A, Entity B"), DO NOT transcribe these names.
    - You must interpret this list as "the specific group of {entity_type}s that 
      qualified through the previous step".
    - Phrasing: Instead of listing names, use phrases like "among the remaining 
      eligible candidates," "within this refined selection," or "relative to the 
      qualifying peer group."

IV. AGENCY IDENTITY
- Invent a plausible governing authority based on the context 
  (e.g., "Board of Agricultural Standards").
\end{lstlisting}

\section{Examples of Symbolic Reasoning Traces}
\label{sec:appendix_traces}

To illustrate the diversity of reasoning structures in \textsc{DocHop}, we provide additional examples of the sampled symbolic reasoning traces in this appendix. 
Each trace is represented as a logical tree, where nodes correspond to parameterized constraints instantiated from rule templates, and edges propagate the valid entity set through successive filtering steps. 
In addition to simple linear chains, the generation process produces multi-branch compositions with Boolean operators (\textsc{AND}/\textsc{OR}), resulting in varied multi-hop reasoning patterns across documents. 
These traces serve as the underlying blueprint that is later verbalized into document narratives and grounded through the semantic reference label.

\begin{figure}[t]
    \centering
    \includegraphics[width=0.95\linewidth]{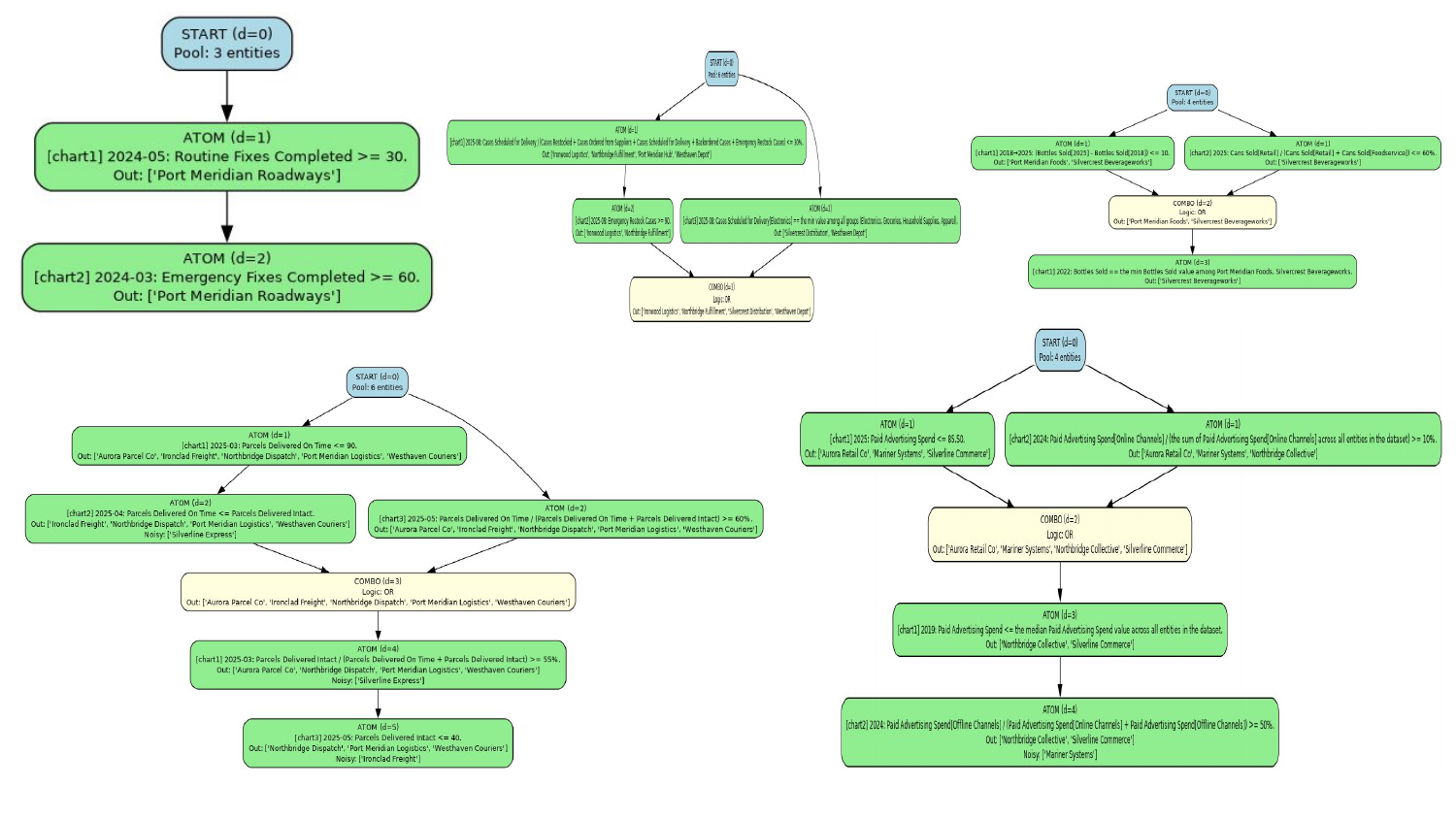}
    \caption{\textbf{Examples of sampled symbolic reasoning traces.} 
    Each trace forms a logical tree of instantiated constraints, ranging from linear multi-hop chains to branched compositions with \textsc{AND}/\textsc{OR} operators.}
    \label{fig:trace_examples}
\end{figure}

Figure~\ref{fig:trace_examples} shows representative reasoning traces with different depths and branching structures.

\section{Atomic Constraint Rule Templates}
\label{sec:appendix_rule_templates}

\paragraph{Atomic Constraint Templates.}
To construct diverse multi-hop reasoning traces, we define a library of atomic constraint templates (``atoms'') that operate over different chart structures.
Each atom specifies a local, parameterized condition grounded in a single chart, such as threshold checks, within-entity comparisons, temporal trends, or extremal selection.
During trace generation, these templates are instantiated by sampling concrete timestamps, metrics, groups, operators, and numeric thresholds from the chart schema, yielding fully specified symbolic constraints.

Importantly, we distinguish between \emph{non-unique} atoms and \emph{unique} atoms.
Non-unique constraints may be satisfied by multiple entities simultaneously (e.g., ``Revenue $\geq$ 70''), enabling gradual filtering across hops.
Unique constraints instead enforce a deterministic extremal selection (e.g., ``the entity with the highest value''), ensuring that the reasoning trace can resolve to a well-defined target set.
Tables~\ref{tab:atom_templates_line}--\ref{tab:atom_templates_groupbar} summarize all atomic templates used in DocHop.

\begin{table*}[t]
\centering
\small
\setlength{\tabcolsep}{6pt}
\renewcommand{\arraystretch}{1.2}
\resizebox{\textwidth}{!}{
\begin{tabular}{l l l}
\toprule
\textbf{Template} & \textbf{Constraint Meaning} & \textbf{Example Instantiation} \\
\midrule

timestamp\_threshold &
Value at a specific timestamp exceeds/falls below a threshold &
``In 2023, Sales $\geq$ 70.'' \\

timestamp\_vs\_across\_entity\_agg &
Value at a timestamp compared against an aggregate across entities &
``In 2022, A's Revenue $\geq$ the mean Revenue of all entities.'' \\

timestamp\_vs\_within\_entity\_time\_agg &
Value at a timestamp compared against an aggregate over time for the same entity &
``In 2021, Profit $\leq$ A's average Profit across all years.'' \\

two\_point\_change\_threshold &
Absolute change between two timestamps exceeds/falls below a threshold &
``From 2020 to 2022, Growth increases by at least 15.'' \\

two\_point\_percentage\_change &
Percentage change between two timestamps satisfies a condition &
``From 2019 to 2023, Output rises by more than 20\%.'' \\

window\_monotone &
Values over a consecutive time window are monotonic &
``From 2021--2024, Engagement is strictly increasing.'' \\

window\_slope\_strict\_monotone &
Slope trend across a window follows strict directional consistency &
``The rate of increase accelerates each year from 2020--2023.'' \\

timestamp\_extreme (unique) &
Entity achieves the maximum/minimum value at a timestamp &
``In 2024, A has the highest Satisfaction.'' \\

window\_sum\_extreme (unique) &
Entity has the largest/smallest cumulative value over a time window &
``Across 2020--2022, B has the lowest total Cost.'' \\

\bottomrule
\end{tabular}
}
\caption{\textbf{Atomic constraint templates for line charts.}
These rules impose temporal thresholds, trends, and extremal conditions over time-series values.}
\label{tab:atom_templates_line}
\end{table*}

\begin{table*}[t]
\centering
\small
\setlength{\tabcolsep}{6pt}
\renewcommand{\arraystretch}{1.2}
\resizebox{\textwidth}{!}{
\begin{tabular}{l l l}
\toprule
\textbf{Template} & \textbf{Constraint Meaning} & \textbf{Example Instantiation} \\
\midrule

metric\_threshold &
A metric value exceeds/falls below a threshold &
``In 2024, Revenue $\geq$ 60.'' \\

row\_metric\_extreme &
One metric is the largest/smallest within an entity's metric profile &
``For A, Profit is the highest among all metrics.'' \\

row\_metric\_vs\_row\_stat &
Metric compared against entity-level statistics (mean/sum) &
``For B, Cost $\leq$ B's average across metrics.'' \\

row\_share\_threshold &
Metric contributes at least a fraction of the entity's total &
``For C, Marketing accounts for at least 40\% of total spend.'' \\

within\_entity\_metric\_compare &
Direct comparison between two metrics of the same entity &
``For A, Revenue $\geq$ Expense.'' \\

metric\_difference\_threshold &
Difference between two metrics exceeds/falls below a threshold &
``For D, Profit minus Cost $\geq$ 20.'' \\

single\_metric\_cross\_entity\_extreme (unique) &
Entity is extremal under one metric across all entities &
``A has the lowest Transaction Fees.'' \\

row\_total\_cross\_entity\_extreme (unique) &
Entity has extremal total across all metrics &
``B has the highest overall expenditure.'' \\

\bottomrule
\end{tabular}
}
\caption{\textbf{Atomic constraint templates for multi-metric bar charts.}
These rules capture within-entity metric comparisons and cross-entity extrema.}
\label{tab:atom_templates_multimetric}
\end{table*}

\begin{table*}[t]
\centering
\small
\setlength{\tabcolsep}{6pt}
\renewcommand{\arraystretch}{1.2}
\resizebox{\textwidth}{!}{
\begin{tabular}{l l l}
\toprule
\textbf{Template} & \textbf{Constraint Meaning} & \textbf{Example Instantiation} \\
\midrule

within\_entity\_group\_compare &
Compare values across groups for the same entity &
``For A, North $\geq$ South.'' \\

within\_entity\_group\_share\_threshold &
A group contributes at least a fraction of entity total &
``For B, Online accounts for at least 50\% of total.'' \\

same\_group\_value\_threshold &
Threshold applied to a specific group value across entities &
``In Retail, Sales $\geq$ 70.'' \\

same\_group\_share\_threshold &
Group share constraint across entities &
``For all entities, APAC contributes less than 30\%.'' \\

same\_group\_vs\_row\_stat &
Group value compared against entity-level statistics &
``For C, Europe $\leq$ C's average across regions.'' \\

group\_difference\_threshold &
Difference between two groups exceeds/falls below threshold &
``For D, East minus West $\geq$ 15.'' \\

row\_group\_extreme &
Within an entity, one group is maximal/minimal &
``For A, North is the largest region.'' \\

same\_group\_value\_extreme (unique) &
Entity achieves extremum in a specific group &
``B has the highest value in APAC.'' \\

same\_group\_share\_extreme (unique) &
Entity has extremal proportional share in a group &
``C has the largest APAC share.'' \\

row\_groups\_sum\_extreme (unique) &
Entity has extremal total summed across groups &
``D has the lowest total output across all regions.'' \\

\bottomrule
\end{tabular}
}
\caption{\textbf{Atomic constraint templates for group bar charts.}
These rules impose within-entity group structure and cross-entity extremal selection.}
\label{tab:atom_templates_groupbar}
\end{table*}

\section{Question Templates}
\label{sec:appendix_question_templates}

Given a synthesized document instance, we construct the final evaluation question by instantiating a task-specific template.
Each template defines a canonical chart question-answering format (e.g., retrieval, counting, aggregation, ranking, hypothetical updates, or fact checking), while leaving key slots---such as the target metric, timestamp, aggregation operator, or additional condition---to be filled from the instance metadata.

Crucially, all templates are grounded on the \emph{semantic reference label} introduced at the end of the document narrative, rather than directly naming entities from the charts.
As a result, questions do not explicitly specify which rows in the chart tables should be queried.
Instead, models must first resolve the labeled entity set implied by the narrative constraints, and then retrieve or aggregate the corresponding numerical evidence from the charts.

Across the six task categories, we design a total of 41 diverse templates.
These templates cover both direct queries over the labeled entities (e.g., retrieving a metric value or counting the number of qualifying entities) and more compositional variants that introduce additional chart-grounded conditions or counterfactual modifications.
This template-based construction enables systematic control over question forms while ensuring that all answers can be computed exactly from the underlying chart tables.
The complete template list, grouped by task type, is provided in the following pages.

\renewcommand{\arraystretch}{0.85}
\begin{table*}[t]
\centering
\scriptsize
\begin{tabular}{|l|l|p{11cm}|}
\hline
\textbf{Template ID} & \textbf{Task} & \textbf{Question Pattern} \\ \hline

\multicolumn{3}{|c|}{\textbf{Task 1: Value Retrieval}} \\ \hline
1.1 & Value Query & What is the \{metric\_desc\} of the \{entity\_type\} receiving the \{semantic\_label\}? \\ \hline
1.2 & Entity Query & Which \{entity\_type\} receives the \{semantic\_label\}? \\ \hline

\multicolumn{3}{|c|}{\textbf{Task 2: Counting}} \\ \hline
2.1 & Count All & How many \{entity\_type\}(s) receive the \{semantic\_label\}? \\ \hline
2.2 & Count w/ Threshold & Among the \{entity\_type\}(s) receiving the \{semantic\_label\}, how many have \{metric\_desc\} \{comparison\} \{threshold\}? \\ \hline

\multicolumn{3}{|c|}{\textbf{Task 3: Numeric Reasoning}} \\ \hline
3.1 & Sum & What is the total \{metric\_desc\} of the \{entity\_type\}(s) receiving the \{semantic\_label\}? \\ \hline
3.2 & Range & What is the difference between the highest and lowest \{metric\_desc\} among the \{entity\_type\}(s) receiving the \{semantic\_label\}? \\ \hline
3.3 & Max & What is the highest \{metric\_desc\} among the \{entity\_type\}(s) receiving the \{semantic\_label\}? \\ \hline
3.4 & Min & What is the lowest \{metric\_desc\} among the \{entity\_type\}(s) receiving the \{semantic\_label\}? \\ \hline
3.5 & Average & What is the average \{metric\_desc\} of the \{entity\_type\}(s) receiving the \{semantic\_label\}? \\ \hline
3.6 & Time Range Sum & What is the total \{metric\} from \{start\_time\} to \{end\_time\} of all the \{entity\_type\}s receiving the \{semantic\_label\}? \\ \hline
3.7 & Time Diff & What is the difference in the total \{metric\} between \{time1\} and \{time2\} of all the \{entity\_type\}s receiving the \{semantic\_label\}? \\ \hline
3.8 & Multi-Metric Sum & What is the total of \{metric1\} and \{metric2\} in \{time\} for all the \{entity\_type\}s receiving the \{semantic\_label\}? \\ \hline
3.9 & Multi-Metric Diff & What is the difference between the total \{metric1\} and total \{metric2\} in \{time\} of all the \{entity\_type\}s receiving the \{semantic\_label\}? \\ \hline
3.10 & Group Sum & What is the combined total \{metric\} across \{group1\} and \{group2\} in \{time\} for all the \{entity\_type\}s receiving the \{semantic\_label\}? \\ \hline
3.11 & Group Diff & What is the difference in total \{metric\} between \{group1\} and \{group2\} in \{time\} of all the \{entity\_type\}s receiving the \{semantic\_label\}? \\ \hline

\multicolumn{3}{|c|}{\textbf{Task 4: Ranking}} \\ \hline
4.1 & Argmax & Among the \{entity\_type\}(s) receiving the \{semantic\_label\}, which one has the highest \{metric\_desc\}? \\ \hline
4.2 & Argmin & Among the \{entity\_type\}(s) receiving the \{semantic\_label\}, which one has the lowest \{metric\_desc\}? \\ \hline
4.3 & K-th Best & Among the \{entity\_type\}(s) receiving the \{semantic\_label\}, which one has the \{k-th\} highest \{metric\_desc\}? \\ \hline

\end{tabular}
\caption{Question Templates for Chart-based QA}
\label{tab:qa_templates}
\end{table*}

\begin{table*}[t]
\centering
\scriptsize
\begin{tabular}{|l|l|p{11cm}|}
\hline
\textbf{Template ID} & \textbf{Task} & \textbf{Question Pattern} \\ \hline

\multicolumn{3}{|c|}{\textbf{Task 5: Hypothetical}} \\ \hline
5.1 & Global Change & If the total \{metric\_desc\} of all the \{entity\_type\}s receiving the \{semantic\_label\} \{increased/decreased\} by \{pct\}\%, what would be the new value? \\ \hline
5.2 & Extremum Change & If the \{entity\_type\} with the \{highest/lowest\} \{metric\_desc\} (among those receiving the \{semantic\_label\}) had their value \{increased/decreased\} by \{pct\}\%, what would be the new total \{metric\_desc\} for all recipients? \\ \hline
5.3 & Add Constraint & Suppose that \{entity\_type\}s that would otherwise qualify for the \{semantic\_label\} are now additionally required to satisfy the following constraint: \{new\_condition\}. \{downstream\_question\} \\ \hline
5.4 & Modify Threshold & Suppose that, for \{entity\_type\}s seeking the \{semantic\_label\}, the threshold defined under \{constraint\_name\} was adjusted from \{old\_threshold\} to \{new\_threshold\}. \{downstream\_question\} \\ \hline
5.5 & Flip Logic & Suppose that, for \{entity\_type\}s seeking the \{semantic\_label\}, the rule in \{section\_name\} was modified to require satisfying all listed conditions simultaneously rather than qualifying through only one. \{downstream\_question\} \\ \hline
5.6 & Swap Dimension & Suppose that, for \{entity\_type\}s seeking the \{semantic\_label\}, the evaluation metric in \{section\_name\} was changed from `\{old\_metric\}' to `\{new\_metric\}'. \{downstream\_question\} \\ \hline

\end{tabular}
\caption{Question Templates for Chart-based QA (Continued - Hypothetical)}
\label{tab:qa_templates_hyp}
\end{table*}

\begin{table*}[t]
\centering
\scriptsize
\begin{tabular}{|l|l|p{11cm}|}
\hline
\textbf{Template ID} & \textbf{Subtype} & \textbf{Question Pattern} \\ \hline

\multicolumn{3}{|c|}{\textbf{Task 6: Fact Checking (Yes/No)}} \\ \hline

\multicolumn{3}{|l|}{\textit{6.1 - Entity Verification}} \\ \hline
6.1.1 & Entity Membership & Is \{entity\} listed as a recipient of the \{semantic\_label\}? (Yes/No) \\ \hline
6.1.2 & Value Match & Did the \{entity\_type\} receiving the \{semantic\_label\} record a \{metric\_desc\} of \{value\}? (Yes/No) \\ \hline

\multicolumn{3}{|l|}{\textit{6.2 - Count Verification}} \\ \hline
6.2.1 & Exact Count & Are there exactly \{count\} \{entity\_type\}(s) receiving the \{semantic\_label\}? (Yes/No) \\ \hline

\multicolumn{3}{|l|}{\textit{6.3 - Numeric Verification}} \\ \hline
6.3.1 & Sum Check & Is the total \{metric\_desc\} of the \{entity\_type\}(s) receiving the \{semantic\_label\} equal to \{value\}? (Yes/No) \\ \hline
6.3.2 & Range Check & Is the difference between the highest and lowest \{metric\_desc\} among the \{entity\_type\}(s) receiving the \{semantic\_label\} equal to \{value\}? (Yes/No) \\ \hline
6.3.3 & Max Check & Is the highest \{metric\_desc\} among the \{entity\_type\}(s) receiving the \{semantic\_label\} equal to \{value\}? (Yes/No) \\ \hline
6.3.4 & Min Check & Is the lowest \{metric\_desc\} among the \{entity\_type\}(s) receiving the \{semantic\_label\} equal to \{value\}? (Yes/No) \\ \hline
6.3.5 & Average Check & Is the average \{metric\_desc\} of the \{entity\_type\}(s) receiving the \{semantic\_label\} equal to \{value\}? (Yes/No) \\ \hline
6.3.6 & Time Range Sum Check & Is the total \{metric\} from \{start\_time\} to \{end\_time\} for all the \{entity\_type\}s receiving the \{semantic\_label\} equal to \{value\}? (Yes/No) \\ \hline
6.3.7 & Time Diff Check & Is the difference in the total \{metric\} between \{time1\} and \{time2\} for all the \{entity\_type\}s receiving the \{semantic\_label\} equal to \{value\}? (Yes/No) \\ \hline
6.3.8 & Multi-Metric Sum Check & Is the total of \{metric1\} and \{metric2\} in \{time\} for all the \{entity\_type\}s receiving the \{semantic\_label\} equal to \{value\}? (Yes/No) \\ \hline
6.3.9 & Multi-Metric Diff Check & Is the difference between the total \{metric1\} and total \{metric2\} in \{time\} of all the \{entity\_type\}s receiving the \{semantic\_label\} equal to \{value\}? (Yes/No) \\ \hline
6.3.10 & Group Sum Check & Is the combined total \{metric\} across \{group1\} and \{group2\} in \{time\} for all the \{entity\_type\}s receiving the \{semantic\_label\} equal to \{value\}? (Yes/No) \\ \hline
6.3.11 & Group Diff Check & Is the difference in total \{metric\} between \{group1\} and \{group2\} in \{time\} of all the \{entity\_type\}s receiving the \{semantic\_label\} equal to \{value\}? (Yes/No) \\ \hline

\multicolumn{3}{|l|}{\textit{6.4 - Ranking Verification}} \\ \hline
6.4.1 & Argmax Check & Does \{entity\} have the highest \{metric\_desc\} among the \{entity\_type\}(s) receiving the \{semantic\_label\}? (Yes/No) \\ \hline
6.4.2 & Argmin Check & Does \{entity\} have the lowest \{metric\_desc\} among the \{entity\_type\}(s) receiving the \{semantic\_label\}? (Yes/No) \\ \hline
6.4.3 & K-th Best Check & Does \{entity\} have the \{k-th\} highest \{metric\_desc\} among the \{entity\_type\}(s) receiving the \{semantic\_label\}? (Yes/No) \\ \hline

\end{tabular}
\caption{Question Templates for Chart-based QA (Continued - Fact Checking)}
\label{tab:qa_templates_fc}
\end{table*}

\section{Evaluation Details}
\label{sec:appendix_eval}
We evaluate model performance using normalized answer accuracy. 
To reduce formatting variance, we append a shared instruction block to every query, requiring models to return the final answer in a dedicated sentence of the form \texttt{``The answer is: <...>''} and to omit units, currency symbols, and percentage signs. 
When the question specifies a rounding scheme, models are instructed to follow it strictly; otherwise, integer answers are expected as integers, and non-integer answers are rounded to two decimal places.

\textbf{Prompting Setup (System Instruction).}
All models are evaluated in a single-image VQA setting where the input is a document-page image containing both narrative text and one or more charts.
We use a unified system-style instruction (\texttt{SHARED\_INSTRUCTION}) appended to the question turn, which enforces (i) a single final answer span, (ii) a canonical numeric formatting policy, and (iii) a fixed output marker (\texttt{The answer is:}) to facilitate robust parsing.
The instruction begins with a short preamble describing the input format (e.g., ``read the text within the image and analyze the chart(s)''), followed by the shared answer-format constraint described above.

\textbf{Answer Parsing and Normalization.}
Given a model response, we extract the predicted answer by searching for the final occurrence of the pattern \texttt{the answer is:} (case-insensitive) and taking the subsequent span; if the marker is missing, we fall back to the last non-empty line.
We further normalize extracted strings by stripping leading \texttt{answer:} prefixes, whitespace, and lightweight formatting tokens (e.g., \texttt{*}, backticks).

\textbf{Correctness Criteria.}
We apply the same two-step procedure to every question, regardless of task type. We first check for a case-insensitive exact match between the normalized prediction and ground truth (e.g., for Yes/No fact-checking answers or exact entity names). If this fails, we fall back to a numeric comparison: we extract the last numeric token from both prediction and ground truth (allowing an optional minus sign and decimals, and ignoring comma separators) and mark the prediction correct if the two values are close under \texttt{math.isclose} with relative tolerance $0.01$ and absolute tolerance $10^{-3}$.

\begin{verbatim}

The image provided is a document page containing both text and chart(s).
Please read the text within the image and analyze the chart(s) to answer
the question.

Your final answer should be a single pure value:
- If the question asks for an entity, use its complete name exactly as shown 
  in the chart
- If the answer is a number: output integers as integers, decimals rounded to
  two places 
  (unless the question specifies otherwise)
- Otherwise, follow the question's instructions for the expected format

Do not include units, currency symbols, or percentage signs in your final answer.

At the end of your response, format the final answer in a separate sentence 
like this:
The answer is: <your_final_answer>

\end{verbatim}

\section{Model Configuration}
\label{sec:model_config}

\renewcommand{\arraystretch}{1.05}
\begin{table*}[!t]
\centering
\footnotesize
\begin{tabular}{|p{4.3cm}|p{2.6cm}|p{6.2cm}|}
\hline
\textbf{Model Name} & \textbf{Category} & \textbf{Hugging Face Checkpoint / API} \\ \hline

\multicolumn{3}{|c|}{\textbf{Reasoning Models}} \\ \hline
GPT-5.2-Reasoning & Proprietary & gpt-5.2-2025-12-11 \\ \hline
GPT-5-mini-Reasoning & Proprietary & gpt-5-mini-2025-08-07 \\ \hline
Gemini-2.5-Pro-Reasoning & Proprietary & gemini-2.5-pro \\ \hline
Gemini-2.5-Flash-Reasoning & Proprietary & gemini-2.5-flash \\ \hline
\multicolumn{3}{|c|}{\textbf{Proprietary Models}} \\ \hline
GPT-5.2 & Proprietary & gpt-5.2-2025-12-11 \\ \hline
Gemini-2.5-Flash & Proprietary & gemini-2.5-flash \\ \hline
Claude-4.5-Haiku & Proprietary & claude-haiku-4-5-20251001 \\ \hline
Claude-4.5-Sonnet & Proprietary & claude-sonnet-4-5-20250929 \\ \hline

\multicolumn{3}{|c|}{\textbf{Open-Source Models}} \\ \hline
LLaVA-Next-LLaMA3-8B & Open-Source & lmms-lab/llama3-llava-next-8b \\ \hline
IDEFICS2-8B & Open-Source & HuggingFaceM4/idefics2-8b \\ \hline
IDEFICS3-LLaMA3-8B & Open-Source & HuggingFaceM4/Idefics3-8B-Llama3 \\ \hline
Qwen-2.5-VL-7B & Open-Source & Qwen/Qwen2.5-VL-7B-Instruct \\ \hline
Qwen-2.5-VL-32B & Open-Source & Qwen/Qwen2.5-VL-32B-Instruct \\ \hline
Qwen-2.5-VL-72B & Open-Source & Qwen/Qwen2.5-VL-72B-Instruct \\ \hline
Qwen3-VL-8B & Open-Source & Qwen/Qwen3-VL-8B-Instruct \\ \hline
Molmo-7B-O-0924 & Open-Source & allenai/Molmo-7B-O-0924 \\ \hline
Molmo-7B-D-0924 & Open-Source & allenai/Molmo-7B-D-0924 \\ \hline
InternVL-3.5-8B & Open-Source & OpenGVLab/InternVL3\_5-8B \\ \hline
InternVL-3.5-30B-A3B & Open-Source & OpenGVLab/InternVL3\_5-30B-A3B \\ \hline
InternVL-3.5-38B & Open-Source & OpenGVLab/InternVL3\_5-38B \\ \hline
Ovis1.6-Gemma2-9B & Open-Source & AIDC-AI/Ovis1.6-Gemma2-9B \\ \hline
\end{tabular}
\caption{Model List and Checkpoints}
\label{tab:model_checkpoints}
\end{table*}

We report the evaluation configurations of all models considered in our experiments.
For proprietary APIs, we specify the exact model versions used at evaluation time.
For open-sourced baselines, we list the corresponding HuggingFace checkpoints.

For Qwen-2.5-VL and Qwen3-VL models, we follow the recommended visual tokenization
settings, using a minimum pixel resolution of $1280 \times 28 \times 28$ and a
maximum resolution of $16384 \times 28 \times 28$.

Since DocHop is reasoning-intensive and often requires multi-step aggregation,
we set the maximum generation length for all models to $2^{14}$ tokens whenever
supported, avoiding premature truncation due to insufficient output budgets.


\end{document}